\documentclass{article} 
\usepackage{iclr2026_conference,times}

\usepackage{amsmath,amsfonts,bm}

\def\eqref#1{equation~\ref{#1}}

\def\1{\bm{1}}

\DeclareMathAlphabet{\mathsfit}{\encodingdefault}{\sfdefault}{m}{sl}
\SetMathAlphabet{\mathsfit}{bold}{\encodingdefault}{\sfdefault}{bx}{n}

\usepackage[none]{hyphenat}
\usepackage{hyperref}
\usepackage{booktabs}
\usepackage{url}
\usepackage{graphicx}

\title{One Operator to Rule Them All? On Boundary-Indexed Operator Families in Neural PDE Solvers}

\author{Lennon J. Shikhman \\
College of Computing, Georgia Institute of Technology \\
Department of Mathematics and Systems Engineering, Florida Institute of Technology \\ \textit{lshikhman3@gatech.edu}
}

\iclrfinalcopy
\begin{document}

\maketitle

\begin{abstract}
Neural PDE solvers are often described as learning solution operators that map problem data to PDE solutions. In this work, we argue that this interpretation is generally incorrect when boundary conditions vary. We show that standard neural operator training implicitly learns a boundary-indexed family of operators, rather than a single boundary-agnostic operator, with the learned mapping fundamentally conditioned on the boundary-condition distribution seen during training. We formalize this perspective by framing operator learning as conditional risk minimization over boundary conditions, which leads to a non-identifiability result outside the support of the training boundary distribution. As a consequence, generalization in forcing terms or resolution does not imply generalization across boundary conditions. We support our theoretical analysis with controlled experiments on the Poisson equation, demonstrating sharp degradation under boundary-condition shifts, cross-distribution failures between distinct boundary ensembles, and convergence to conditional expectations when boundary information is removed. Our results clarify a core limitation of current neural PDE solvers and highlight the need for explicit boundary-aware modeling in the pursuit of foundation models for PDEs.
\end{abstract}

\section{Introduction}
Recent progress in neural methods for partial differential equations (PDEs) has led to models that aim to directly approximate mappings from problem specifications to PDE solutions. Neural operators, physics-informed networks, and related architectures are frequently described as learning solution operators for entire PDE families, enabling rapid evaluation across varying inputs such as forcing functions, coefficients, or discretizations. This viewpoint has fueled growing interest in developing scalable, reusable neural PDE solvers as potential foundation models for scientific computing.

From the perspective of classical PDE theory, however, a solution operator is not defined by the differential equation alone. Boundary conditions are essential to well-posedness and fundamentally determine the mapping from problem data to solutions. In many learning-based approaches, boundary conditions vary across training instances but are incorporated implicitly, for example through boundary padding, auxiliary input channels, or fixed encodings. This raises a basic but unresolved question: when boundary conditions are not fixed, what object is a neural PDE solver actually approximating?

In this paper, we contend that standard neural PDE solvers do not learn a single operator that is invariant to boundary conditions. Rather, they learn a \text{family of operators indexed by boundary conditions}, with the learned mapping intrinsically tied to the boundary-condition distribution present in the training data. This interpretation exposes a structural limitation of current training paradigms. Because learning is driven by empirical risk minimization, the model is only constrained on the subset of boundary conditions it observes, leaving its behavior under boundary shifts or extrapolation effectively unconstrained.

We develop this argument by explicitly casting neural operator learning as conditional risk minimization with respect to boundary conditions. Within this framework, breakdowns in boundary-condition generalization arise naturally and do not depend on architectural deficiencies or optimization failures. In particular, robustness to changes in forcing terms or spatial resolution does not guarantee robustness to changes in boundary conditions, even when the underlying PDE operator is unchanged. Furthermore, when boundary information is omitted or weakly represented, learned models tend to approximate conditional averages rather than well-defined PDE solution maps.

We illustrate these phenomena through a series of controlled numerical experiments on the Poisson equation with mixed boundary conditions. The experiments reveal pronounced performance degradation under boundary-condition shifts, systematic failures when training and testing boundary distributions differ, and convergence toward conditional expectations when boundary inputs are removed. Together, these results provide empirical support for the theoretical characterization.

Several prior works have documented sensitivity to boundary-condition variability and proposed boundary-aware architectures or constraints to mitigate such effects. Our contribution is not to introduce a new architecture, but to provide a learning-theoretic framing that interprets these empirical behaviors through conditional risk minimization and non-identifiability under boundary distribution shift. This perspective clarifies what object is being learned when boundary conditions vary and sharpens the interpretation of operator-level generalization claims.

\paragraph{Contributions.}
This work makes the following contributions:
\begin{itemize}
    \item We introduce a conceptual framework that interprets neural PDE solvers as learning boundary-indexed operator families rather than boundary-invariant solution operators.
    \item We identify a non-identifiability phenomenon induced by empirical risk minimization, showing that operator behavior outside the training boundary distribution is not determined.
    \item We validate the theoretical predictions with controlled experiments that isolate the effects of boundary-condition variation.
\end{itemize}

\section{Problem Setup and Notation}
\label{sec:setup}

We investigate learning-based approximations of solution mappings for elliptic partial differential equations under varying boundary conditions. To isolate the role of boundary conditions in a controlled setting, we focus on a single canonical problem where the underlying differential operator is fixed while boundary specifications vary across samples.

\paragraph{Differential equation and domain.}
Let $\Omega = [0,1]^2 \subset \mathbb{R}^2$ denote a bounded rectangular domain. For a given forcing field $f : \Omega \rightarrow \mathbb{R}$, we consider the Poisson equation
\begin{equation}
- \Delta u(x,y) = f(x,y), \quad (x,y) \in \Omega,
\end{equation}
supplemented by mixed boundary conditions imposed on the boundary $\partial \Omega$.

\paragraph{Boundary specification.}
Boundary conditions are prescribed independently on each edge of the domain. Specifically, we impose Dirichlet conditions on the left and bottom boundaries and Neumann conditions on the right and top boundaries:
\begin{align}
u(0,y) &= g_L(y), & y \in [0,1], \\
u(x,0) &= g_B(x), & x \in [0,1], \\
\partial_x u(1,y) &= h_R(y), & y \in [0,1], \\
\partial_y u(x,1) &= h_T(x), & x \in [0,1].
\end{align}
The tuple
\[
\mathcal{B} = (g_L, g_B, h_R, h_T)
\]
collects all boundary data required to define a single well-posed problem instance.

In our experiments, the boundary functions are drawn from parameterized families of smooth functions generated via truncated Fourier expansions. By adjusting parameters such as bandwidth, amplitude, and mean offset, we construct distinct boundary-condition distributions while leaving the interior forcing distribution unchanged.

\paragraph{Solution mappings.}
When the boundary data $\mathcal{B}$ are held fixed, the Poisson problem induces a deterministic solution operator
\begin{equation}
\mathcal{S}_{\mathcal{B}} : f \mapsto u.
\end{equation}
When boundary conditions vary across samples, it is more appropriate to view the problem in terms of the joint mapping
\begin{equation}
\mathcal{S} : (f, \mathcal{B}) \mapsto u,
\end{equation}
which assigns a solution to each combination of forcing and boundary specification. Learning-based PDE solvers seek to approximate this joint map from finite data.

\paragraph{Data generation.}
Training and evaluation data are generated synthetically. For each sample, a forcing function $f_i$ is sampled from a fixed distribution of smooth functions on $\Omega$. Boundary data $\mathcal{B}_i$ are sampled independently from a boundary-condition distribution $\mu_{\mathcal{B}}$. The corresponding solution $u_i$ is computed numerically using a finite-difference discretization combined with an iterative Jacobi solver applied on a uniform grid.

We consider multiple choices of $\mu_{\mathcal{B}}$ that differ only in the statistical properties of the boundary functions, such as shifts in mean value or changes in smoothness, while keeping the differential operator and forcing distribution fixed. This design enables a direct examination of how learned solution mappings depend on the boundary-condition distribution observed during training.

\paragraph{Neural solver inputs.}
Neural PDE solvers operate on discretized fields defined on a regular grid. Model inputs consist of the discretized forcing field and spatial coordinate channels. When boundary information is provided, additional channels encode boundary values along with binary masks identifying boundary locations. For comparison, we also consider boundary-ablated models that receive only the forcing and coordinate information, thereby excluding explicit access to boundary conditions.

\section{What Is Actually Being Learned}
\label{sec:what-is-learned}

Neural PDE solvers are often interpreted as approximating solution operators associated with a fixed differential equation. This interpretation becomes ambiguous when boundary conditions vary across training samples, since the PDE no longer defines a single mapping from problem data to solutions.

\paragraph{Learning as conditional prediction.}
In our setting, training data consist of samples $(f_i, \mathcal{B}_i, u_i)$ drawn from a joint distribution over forcing functions and boundary conditions. Standard training minimizes empirical risk of the form
\begin{equation}
\min_{\theta}\; \mathbb{E}_{(f,\mathcal{B})\sim\mu}\!\left[\ell\!\left(\hat{\mathcal{S}}_{\theta}(f,\mathcal{B}),\,\mathcal{S}(f,\mathcal{B})\right)\right],
\end{equation}
which corresponds to learning a conditional predictor of $u$ given $(f,\mathcal{B})$. As a result, the learned model should be viewed as approximating a collection of solution maps indexed by boundary conditions, rather than a boundary-invariant operator.

\paragraph{Dependence on the boundary distribution.}
Because empirical risk minimization constrains the model only on the support of the training distribution, the learned mapping is well-defined only for boundary conditions drawn from the training boundary distribution $\mu_{\mathcal{B}}$. For boundary conditions outside this support, the training objective imposes no constraints, and multiple extensions of the learned mapping are equally valid. Consequently, two models trained under different boundary-condition distributions may converge to different solution mappings, even when the underlying PDE and forcing distribution are identical.

\paragraph{ERM interpretation.}
Under squared loss, empirical risk minimization yields the conditional expectation of the target given the observed inputs. Our contribution is to interpret this standard result in the operator-learning setting where boundary variability may be unobserved or weakly encoded. Within the class of ERM-trained Fourier Neural Operator models studied here, this behavior is therefore not attributable to insufficient capacity or optimization instability. When boundary information is removed or poorly represented, ERM favors predictors that approximate conditional averages over the unobserved boundary variables,
\begin{equation}
\hat{u}(f) \approx \mathbb{E}_{\mathcal{B}\sim\mu_{\mathcal{B}}}\!\left[\mathcal{S}(f,\mathcal{B}) \mid f\right],
\end{equation}
which generally does not correspond to the solution operator for any fixed boundary condition.

This perspective clarifies why robustness along one axis of variability does not necessarily imply robustness along another, and motivates viewing neural PDE solvers as learning families of operators parameterized by boundary conditions.

\section{Non-Identifiability Induced by Boundary Distribution Shift}
\label{sec:nonidentifiability}

When boundary conditions vary across training samples, the solution map associated with a fixed differential operator is no longer uniquely determined by the training objective alone. In this setting, learning-based PDE solvers are tasked with approximating a mapping that is only partially observed, with key degrees of freedom controlled by the boundary-condition distribution encountered during training.

We use the term \emph{non-identifiability} to describe this phenomenon. Informally, a learned solution map is non-identifiable if multiple distinct mappings achieve the same empirical risk under the training distribution but behave differently under boundary conditions not observed during training. This notion is distinct from approximation error: even in the limit of infinite model capacity and perfect optimization, the training objective does not select a unique extension of the learned mapping beyond the support of the boundary-condition distribution.

This behavior follows directly from the structure of empirical risk minimization. During training, errors are penalized only on samples drawn from the joint distribution over forcings and boundary conditions. For boundary conditions that occur with nonzero probability under this distribution, the learned model is constrained to approximate the corresponding solutions accurately. For boundary conditions outside this support, however, the objective provides no information, and the learned mapping may vary arbitrarily without affecting training loss.

From this perspective, boundary-condition generalization is fundamentally different from generalization in forcing terms or spatial resolution. Variations in forcing are explicitly represented in the input and are typically sampled densely during training, whereas boundary conditions may occupy a low-dimensional or sparsely sampled subspace. As a result, robustness to unseen forcing functions does not imply robustness to unseen boundary specifications, even when the underlying PDE remains unchanged.

The implications of this non-identifiability are directly reflected in our empirical results. Models trained under one boundary-condition distribution exhibit sharp performance degradation when evaluated under a shifted boundary distribution, despite unchanged forcing statistics and identical differential operators. Within the class of ERM-trained Fourier Neural Operator models studied here, this failure is not attributable to insufficient capacity or optimization instability, as the same architecture performs well in-distribution.

Finally, non-identifiability also explains the behavior observed when boundary information is removed or weakly encoded. In this case, the learning objective cannot condition on boundary conditions at all, and empirical risk minimization instead favors predictors that average over the unobserved boundary variability. The resulting mapping corresponds to a conditional expectation over boundary conditions rather than a valid solution operator for any fixed boundary specification.

Taken together, these considerations show that non-identifiability under boundary distribution shift arises naturally under empirical risk minimization in the class of Fourier Neural Operator models studied here, rather than from insufficient capacity or optimization instability.

\section{Implications for Generalization}
\label{sec:implications}

The perspective developed above has several immediate implications for how generalization in neural PDE solvers should be interpreted and evaluated. In particular, it clarifies why success along certain axes of variation does not translate to robustness under boundary-condition shift, and why common evaluation protocols may substantially overestimate operator-level generalization.

\paragraph{Forcing generalization versus boundary generalization.}
Generalization in forcing terms is often taken as evidence that a neural PDE solver has learned the underlying solution operator. However, forcing functions are typically sampled densely during training and enter the problem as interior data. By contrast, boundary conditions define constraints on the solution space and may vary over a much lower-dimensional and sparsely sampled space. As a result, robustness along one axis of variability does not necessarily imply robustness along another. While we do not explicitly evaluate forcing-distribution shift in this work, boundary variability occupies a distinct structural role under empirical risk minimization, and therefore generalization across forcing and boundary conditions need not coincide.

\paragraph{Resolution robustness does not imply boundary robustness.}
Similarly, robustness to changes in spatial resolution is often cited as a hallmark of operator learning. While resolution generalization reflects stability with respect to discretization, it does not address variability in the boundary specification. A model that performs consistently across grids may still fail catastrophically when boundary conditions shift, even if the continuous PDE and forcing distribution remain unchanged. Resolution robustness therefore probes numerical consistency rather than invariance to problem-defining constraints.

\paragraph{Reinterpreting prior empirical results.}
These observations suggest a reinterpretation of prior empirical successes reported for neural PDE solvers. In many benchmark settings, boundary conditions are fixed or vary within a narrow distribution, implicitly collapsing the problem to learning a single operator or a small family thereof. Under such conditions, strong empirical performance is expected and does not contradict the limitations identified here. Conversely, failures under boundary-condition shift should not be viewed as anomalous but as predictable consequences of the learning objective and data distribution.

\paragraph{Implications for model design and evaluation.}
Taken together, these considerations indicate that meaningful claims of operator generalization require explicit treatment of boundary-condition variability. Without boundary-aware representations or invariance mechanisms, scaling model capacity or training data alone is insufficient to ensure robustness beyond the training boundary distribution. Evaluations that do not probe boundary-condition shift risk conflating in-distribution interpolation with genuine operator-level generalization.

This analysis motivates treating boundary conditions as first-class objects in the design and assessment of neural PDE solvers, particularly in the context of developing reusable or foundation-level models for scientific computing.

\section{Experimental Verification}
\label{sec:experiments}

We empirically validate the theoretical perspective developed above through controlled experiments on the two-dimensional Poisson equation. The experiments are specifically designed to isolate the effect of boundary-condition variability by holding the differential operator, forcing distribution, discretization, and resolution fixed, while varying only the boundary-condition distribution. We evaluate both boundary-aware and boundary-ablated Fourier Neural Operator models under in-distribution, cross-distribution, and extrapolation settings to directly test the claims about boundary-indexed learning and non-identifiability.

\subsection{Experimental Setup}

\paragraph{PDE and discretization.}
All experiments consider the two-dimensional Poisson equation on the unit square $\Omega = [0,1]^2$ with mixed boundary conditions. The equation is discretized on a uniform $64 \times 64$ grid using a standard five-point finite-difference stencil. Ground-truth solutions are computed using a batched Jacobi iterative solver with a fixed number of iterations.

\paragraph{Boundary-condition distributions.}
Boundary conditions are parameterized by smooth functions sampled from truncated Fourier series. For both distributions, Dirichlet and Neumann boundary functions are generated with Fourier bandwidth $K=6$. 

In $\mu_{B_0}$, Dirichlet boundary functions have amplitude $1.0$ with zero mean shift, while Neumann boundary functions have amplitude $0.5$ with zero mean shift. In $\mu_{B_1}$, the Dirichlet amplitude remains $1.0$ but includes a mean shift of $0.6$, while Neumann boundary functions have amplitude $0.8$ with a mean shift of $0.2$. The functional form and bandwidth are otherwise identical across distributions.

Forcing functions are sampled independently from a truncated Fourier expansion with bandwidth $K=6$ and amplitude $3.0$, and this forcing distribution is held fixed across all experiments.

\begin{figure}[h]
    \centering
    \includegraphics[width=0.48\linewidth]{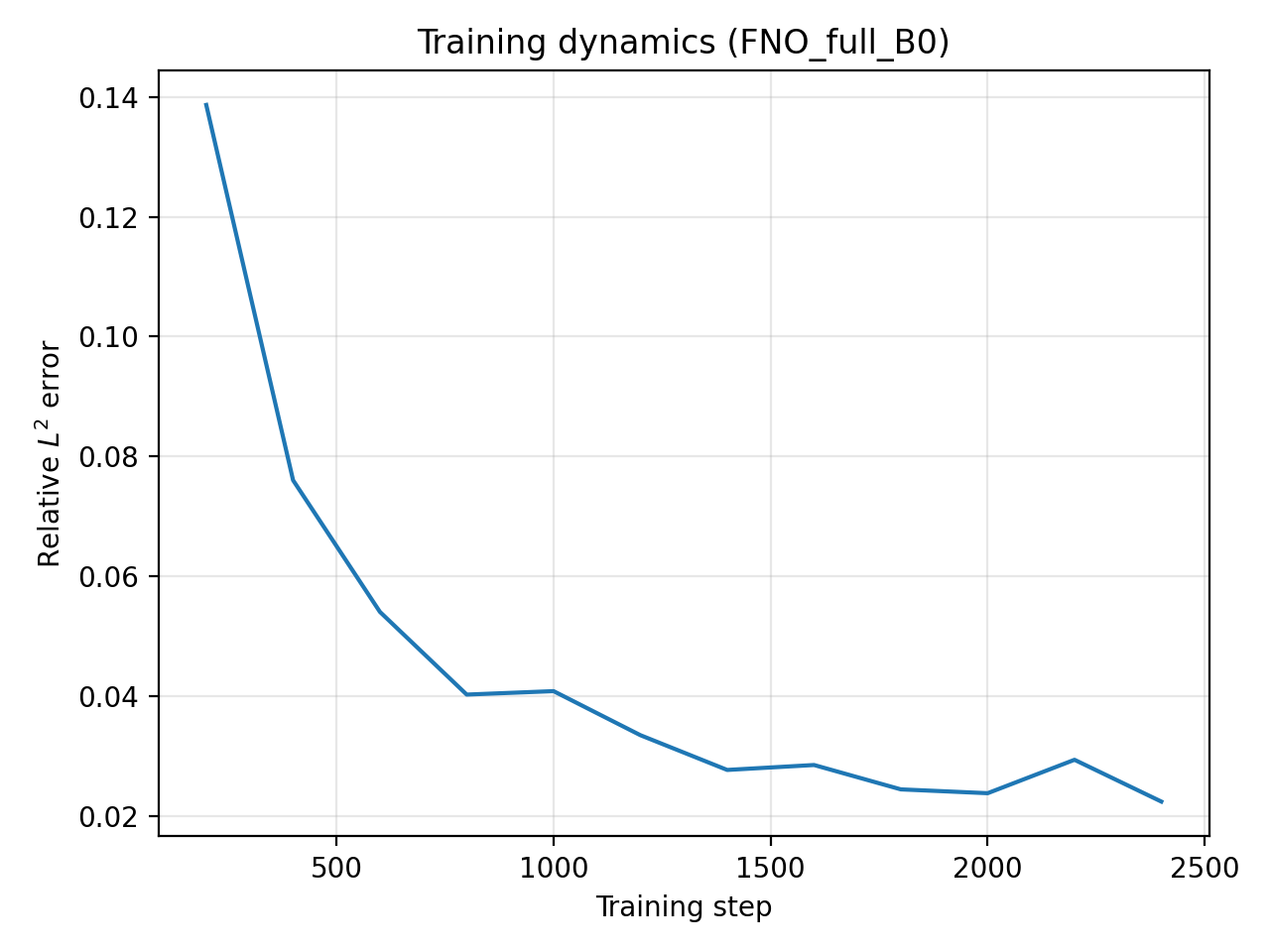}
    \hfill
    \includegraphics[width=0.48\linewidth]{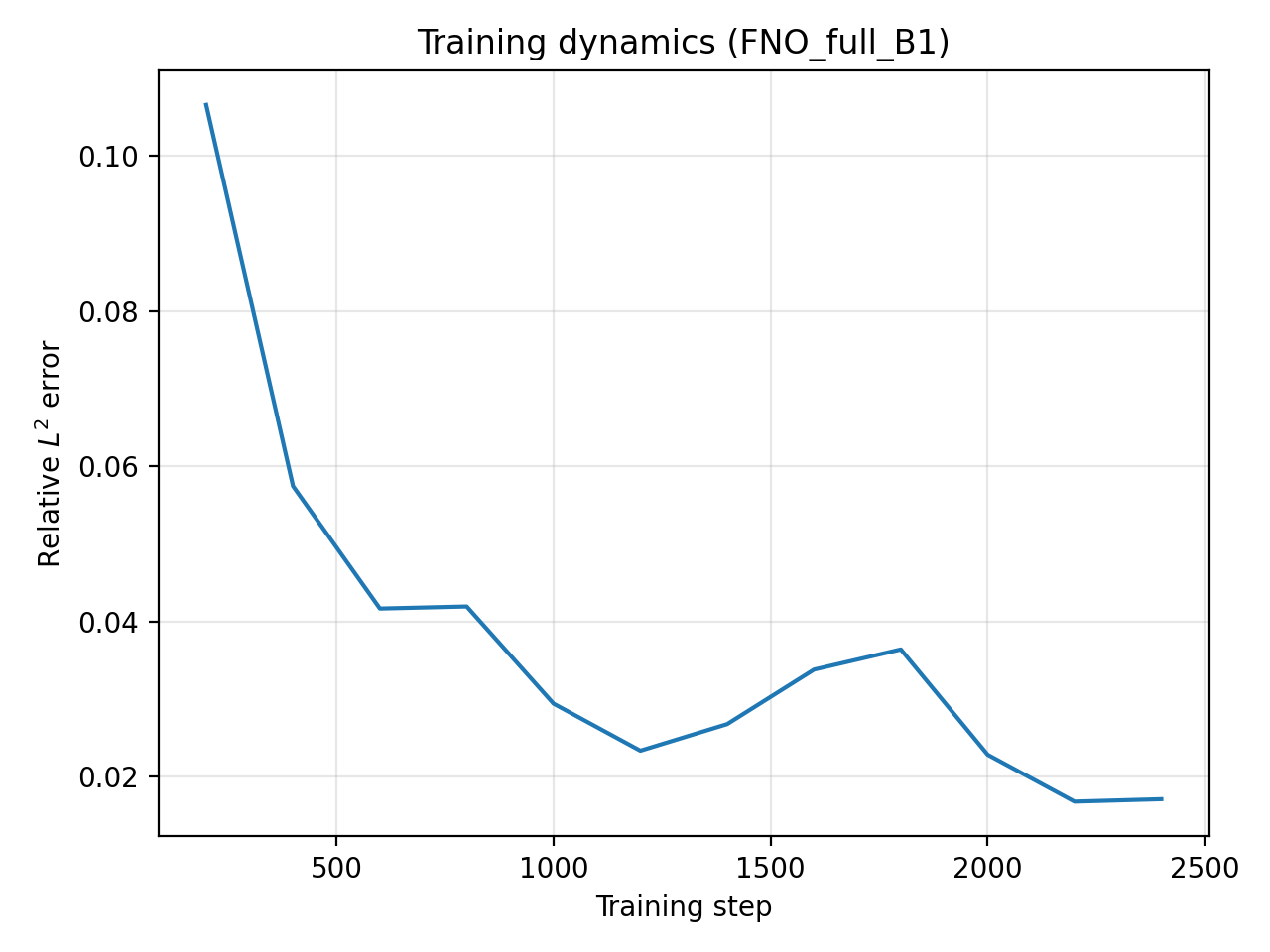}
    \caption{Training dynamics for boundary-aware Fourier Neural Operators trained on $\mu_{B_0}$ (left) and $\mu_{B_1}$ (right). Relative $L^2$ error decreases smoothly and stabilizes in both cases, indicating that subsequent generalization failures are not due to optimization instability or lack of convergence.}
    \label{fig:training_curves}
\end{figure}

\paragraph{Models and training.}
We evaluate a Fourier Neural Operator (FNO) architecture trained under three settings: (i) an FNO trained on $\mu_{B_0}$ with explicit boundary-condition channels, (ii) an FNO trained on $\mu_{B_1}$ with explicit boundary-condition channels, and (iii) a boundary-ablated FNO trained on $\mu_{B_0}$ that receives only the forcing field and spatial coordinates.

All models are trained using the Adam optimizer with learning rate $8\times 10^{-4}$ and mean-squared error loss for $2500$ gradient steps with batch size $12$. Experiments are conducted on a $64\times 64$ grid. Training data are generated using $220$ Jacobi iterations per sample, while evaluation data use $320$ iterations. The random seed is fixed to $7$ for reproducibility. Variability across random initializations is not explored in this study and remains future work.

\paragraph{Error metric.}
Relative $L^2$ error is computed as
\[
\frac{\sqrt{\frac{1}{HW}\sum_{i,j}(u_{\text{pred}}(i,j)-u(i,j))^2}}
{\sqrt{\frac{1}{HW}\sum_{i,j}u(i,j)^2}},
\]
corresponding to the ratio of root-mean-square error to root-mean-square ground-truth magnitude over the spatial grid.

\subsection{Cross-Distribution Generalization}

We first examine generalization under pure boundary-condition distribution shift. Each boundary-aware model is evaluated both in-distribution and on the alternate boundary distribution, while keeping the PDE, forcing distribution, and resolution unchanged. In-distribution results are computed on independently sampled evaluation batches drawn from the same boundary distribution as training.
\begin{table}[h]
\centering
\caption{Cross-distribution generalization under boundary-condition shift. Relative $L^2$ errors are reported as mean $\pm$ standard deviation over evaluation batches.}
\label{tab:cross_dist}
\begin{tabular}{lcc}
\toprule
\textbf{Model} & \textbf{Test on $\mu_{B_0}$} & \textbf{Test on $\mu_{B_1}$} \\
\midrule
FNO trained on $\mu_{B_0}$ & $0.078 \pm 0.005$ & $0.489 \pm 0.022$ \\
FNO trained on $\mu_{B_1}$ & $0.601 \pm 0.036$ & $0.102 \pm 0.003$ \\
FNO (no BC channels)       & $0.999 \pm 0.001$ & $1.001 \pm 0.001$ \\
\bottomrule
\end{tabular}
\end{table}

As shown in Table~\ref{tab:cross_dist}, each boundary-aware model performs well only on the boundary distribution it was trained on, while exhibiting severe degradation under boundary-condition shift. In contrast, the boundary-ablated model fails uniformly across both distributions. These results demonstrate that the learned solution mappings are strongly indexed by the training boundary-condition distribution, rather than determined solely by the underlying PDE.

\subsection{Boundary Extrapolation}

We next evaluate robustness to boundary extrapolation by modifying boundary conditions beyond the support of the training distribution while keeping all other factors fixed.

\paragraph{Dirichlet mean shift.}
We apply additive shifts to the Dirichlet boundary values at test time and measure performance as a function of the shift magnitude. Error increases smoothly and symmetrically as boundary conditions move away from the training distribution, indicating continuous degradation rather than abrupt failure. We sweep mean shifts $\delta \in \{-1.0,-0.5,-0.25,0,0.25,0.5,1.0\}$ applied to the Dirichlet boundary mean relative to the training distribution.

\begin{figure}[h]
    \centering
    \includegraphics[width=0.7\linewidth]{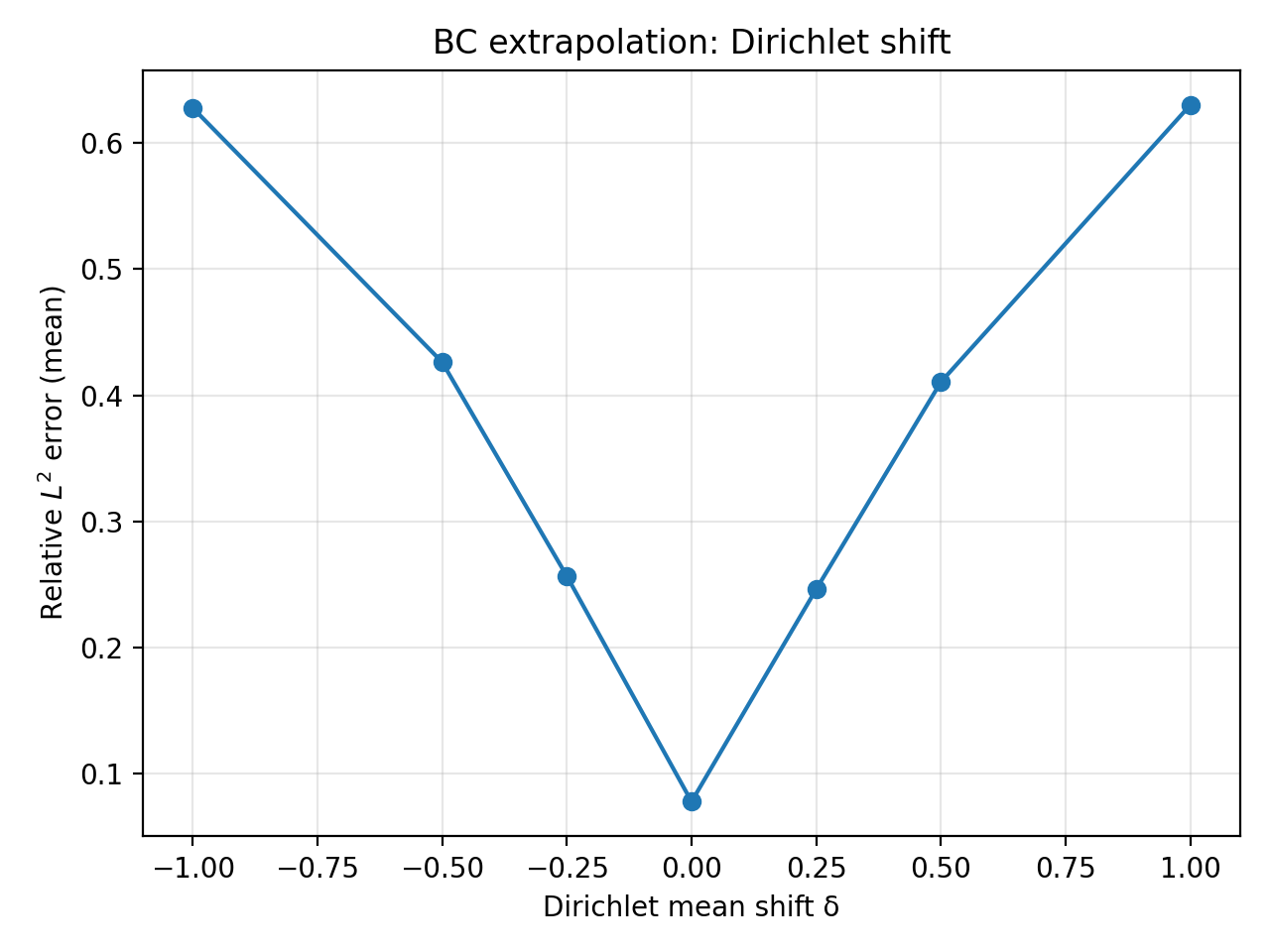}
    \caption{Boundary extrapolation via Dirichlet mean shifts. Relative $L^2$ error increases smoothly and symmetrically as boundary conditions move away from the training distribution, despite unchanged forcing distribution, resolution, and PDE operator.}
    \label{fig:bc_shift}
\end{figure}

\paragraph{Frequency extrapolation.}
We also increase the Fourier bandwidth of the Dirichlet boundary functions beyond the training range. Performance degrades monotonically as higher-frequency boundary components are introduced, despite unchanged forcing statistics and resolution. We increase the Dirichlet bandwidth from the training value $K=6$ to $K \in \{6,8,10,12\}$.

Together, these results show that even modest extrapolation in boundary-condition space leads to substantial error growth, reinforcing the view that learned operators are constrained to the support of the training boundary distribution.

\begin{figure}[h]
    \centering
    \includegraphics[width=0.7\linewidth]{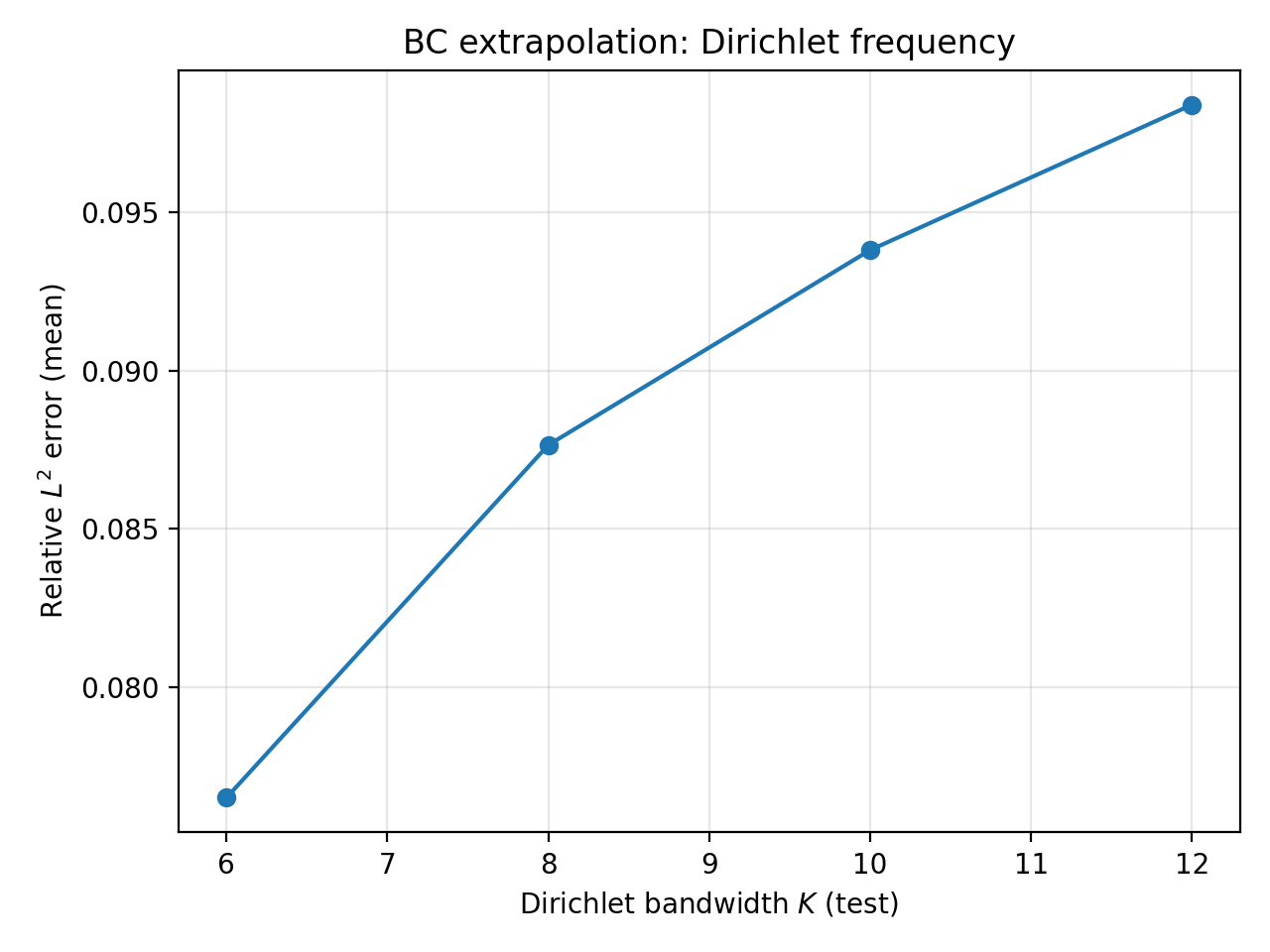}
    \caption{Boundary extrapolation via increased Dirichlet bandwidth. Performance degrades monotonically as higher-frequency boundary components are introduced beyond the training support.}
    \label{fig:bc_freq}
\end{figure}

\subsection{Boundary Ablation and Conditional Expectation}

Finally, we analyze the behavior of models that do not receive explicit boundary-condition information. The boundary-ablated FNO fails to learn a meaningful solution mapping, achieving relative $L^2$ error close to one regardless of the boundary distribution.

To further interpret this behavior, we fix a single forcing function and compare the model output to the empirical average of solutions obtained by sampling boundary conditions from $\mu_{B_0}$. The boundary-ablated model closely matches this conditional average, indicating that empirical risk minimization drives the model toward a conditional expectation over boundary conditions when boundary information is unavailable.

This experiment provides direct empirical evidence for the learning-theoretic interpretation developed in Sections~\ref{sec:what-is-learned} and~\ref{sec:nonidentifiability}, and confirms that boundary ablation leads to averaging behavior rather than recovery of a valid solution operator.

\begin{figure}[h]
    \centering
    \includegraphics[width=0.9\linewidth]{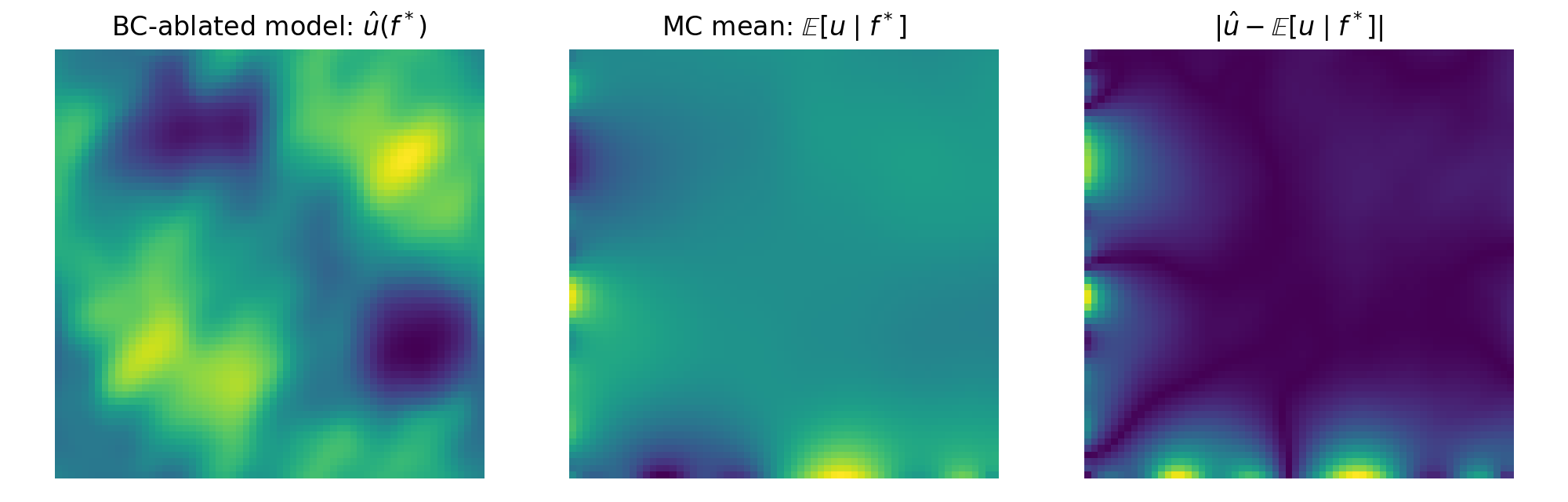}
    \caption{Conditional expectation behavior under boundary ablation. For a fixed forcing $f^*$, the boundary-ablated model output (left) closely matches the Monte Carlo estimate of $\mathbb{E}[u \mid f^*]$ (center), with the absolute difference shown on the right. This provides empirical evidence that training via empirical risk minimization leads to conditional averaging when boundary information is unavailable.}

    \label{fig:condexp}
\end{figure}

\section{Discussion and Implications for Foundation Models}
\label{sec:discussion}

The results of this study have direct implications for how neural PDE solvers should be interpreted, scaled, and evaluated in the context of foundation models for scientific computing. In particular, they clarify which aspects of generalization are structurally achievable under standard training objectives and which are not.

\paragraph{Boundary awareness as a structural requirement.}
Our findings indicate that boundary conditions cannot be treated as incidental inputs when learning solution operators. Instead, they define essential constraints that index distinct solution maps. Models trained without explicit mechanisms to represent or reason over boundary variability inevitably learn mappings that are tied to the boundary-condition distribution observed during training. From this perspective, boundary awareness is not an architectural enhancement but a structural requirement for operator-level generalization.

\paragraph{Limits of scaling under empirical risk minimization.}
The observed failures under boundary-condition shift are not attributable to insufficient data, model capacity, or optimization instability. Rather, they arise from the interaction between empirical risk minimization and variability in boundary conditions. Scaling model size or dataset volume within a fixed boundary distribution improves in-distribution performance but does not resolve non-identifiability outside the support of that distribution. Consequently, scaling alone is insufficient to produce boundary-agnostic solution operators.

\paragraph{Implications for PDE foundation models.}
For foundation models aimed at learning reusable PDE solvers across domains and boundary regimes, these results highlight a critical challenge. Without explicit treatment of boundary conditions, such as invariant representations, conditional operator decompositions, or structured boundary encodings, learned models should be expected to generalize only within the boundary distributions they are trained on. Evaluation protocols that fix or narrowly vary boundary conditions therefore risk overstating the scope of learned operator generalization. These considerations complement recent studies of scaling and transfer behavior in scientific foundation models (e.g., Subramanian et al., 2023), which emphasize the importance of distributional coverage across physical parameters during pre-training and transfer.

Taken together, our analysis suggests that progress toward genuine foundation models for PDEs will require training objectives and representations that address boundary-condition variability as a first-class modeling concern, rather than relying on scale or architectural expressiveness alone.

\section{Limitations and Future Work}
\label{sec:limitations}

This work is intentionally scoped to isolate the effects of boundary-condition variability in a controlled setting. As such, several limitations point to natural directions for future research.

\paragraph{PDE classes.}
Our experiments focus on a single elliptic PDE with smooth solutions and mixed boundary conditions. While the conceptual arguments are not specific to the Poisson equation, extending the empirical analysis to other PDE classes—such as parabolic, hyperbolic, or nonlinear systems—would further clarify how boundary-induced non-identifiability manifests across different dynamical regimes.

\paragraph{Time-dependent systems.}
We restrict attention to steady-state problems. In time-dependent PDEs, boundary conditions may interact with temporal evolution and initial conditions in more complex ways. Investigating whether analogous boundary-indexed behavior arises in neural solvers for evolutionary systems remains an important direction.

\paragraph{Theoretical extensions.}
Our analysis adopts a learning-theoretic perspective rather than providing formal guarantees. Developing more explicit theoretical characterizations of identifiability under boundary variability, as well as designing training objectives or representations that mitigate these effects, are promising avenues for future work.

Overall, these limitations reflect deliberate design choices rather than deficiencies of the proposed analysis, and highlight opportunities for extending the framework developed here.

\section*{Acknowledgments}

\paragraph{Computational Resources.}
The author gratefully acknowledges Dell Technologies, and in particular the Dell Pro Precision division, for providing computational resources that supported the experiments in this work. All experiments were conducted on a Dell Pro Max T2 workstation equipped with an Intel Core Ultra 9 285K processor, 128 GB of DDR5 ECC memory, and an NVIDIA RTX PRO 6000 Blackwell GPU. The views and conclusions expressed herein are those of the author and do not necessarily reflect the views of Dell Technologies.

\paragraph{Reproducibility.}
For reproducibility, the full experimental code, data-generation scripts, and training configurations used in this study are publicly available at
\url{https://github.com/lennonshikhman/boundary-indexed-neural-pde}.

\nocite{*}
\bibliography{iclr2026_conference}

@misc{li2023physicsinformedneuraloperatorlearning,
      title={Physics-Informed Neural Operator for Learning Partial Differential Equations}, 
      author={Zongyi Li and Hongkai Zheng and Nikola Kovachki and David Jin and Haoxuan Chen and Burigede Liu and Kamyar Azizzadenesheli and Anima Anandkumar},
      year={2023},
      eprint={2111.03794},
      archivePrefix={arXiv},
      primaryClass={cs.LG},
      url={https://arxiv.org/abs/2111.03794}, 
}

@misc{li2021fourierneuraloperatorparametric,
      title={Fourier Neural Operator for Parametric Partial Differential Equations}, 
      author={Zongyi Li and Nikola Kovachki and Kamyar Azizzadenesheli and Burigede Liu and Kaushik Bhattacharya and Andrew Stuart and Anima Anandkumar},
      year={2021},
      eprint={2010.08895},
      archivePrefix={arXiv},
      primaryClass={cs.LG},
      url={https://arxiv.org/abs/2010.08895}, 
}

@misc{li2020neuraloperatorgraphkernel,
      title={Neural Operator: Graph Kernel Network for Partial Differential Equations}, 
      author={Zongyi Li and Nikola Kovachki and Kamyar Azizzadenesheli and Burigede Liu and Kaushik Bhattacharya and Andrew Stuart and Anima Anandkumar},
      year={2020},
      eprint={2003.03485},
      archivePrefix={arXiv},
      primaryClass={cs.LG},
      url={https://arxiv.org/abs/2003.03485}, 
}

@book{evans10,
  address = {Providence, R.I.},
  author = {Evans, Lawrence C.},
  isbn = {9780821849743 0821849743},
  publisher = {American Mathematical Society},
  refid = {465190110},
  title = {Partial differential equations},
  year = 2010
}

@article{Lu_2021,
   title={Learning nonlinear operators via DeepONet based on the universal approximation theorem of operators},
   volume={3},
   ISSN={2522-5839},
   url={http://dx.doi.org/10.1038/s42256-021-00302-5},
   DOI={10.1038/s42256-021-00302-5},
   number={3},
   journal={Nature Machine Intelligence},
   publisher={Springer Science and Business Media LLC},
   author={Lu, Lu and Jin, Pengzhan and Pang, Guofei and Zhang, Zhongqiang and Karniadakis, George Em},
   year={2021},
   month=mar, pages={218–229} 
}

@article{JMLR:v24:21-1524,
  author  = {Nikola Kovachki and Zongyi Li and Burigede Liu and Kamyar Azizzadenesheli and Kaushik Bhattacharya and Andrew Stuart and Anima Anandkumar},
  title   = {Neural Operator: Learning Maps Between Function Spaces With Applications to PDEs},
  journal = {Journal of Machine Learning Research},
  year    = {2023},
  volume  = {24},
  number  = {89},
  pages   = {1--97},
  url     = {http://jmlr.org/papers/v24/21-1524.html}
}

@misc{subramanian2023foundationmodelsscientificmachine,
      title={Towards Foundation Models for Scientific Machine Learning: Characterizing Scaling and Transfer Behavior}, 
      author={Shashank Subramanian and Peter Harrington and Kurt Keutzer and Wahid Bhimji and Dmitriy Morozov and Michael Mahoney and Amir Gholami},
      year={2023},
      eprint={2306.00258},
      archivePrefix={arXiv},
      primaryClass={cs.LG},
      url={https://arxiv.org/abs/2306.00258}, 
}

@misc{kovachki2024operatorlearningalgorithmsanalysis,
      title={Operator Learning: Algorithms and Analysis}, 
      author={Nikola B. Kovachki and Samuel Lanthaler and Andrew M. Stuart},
      year={2024},
      eprint={2402.15715},
      archivePrefix={arXiv},
      primaryClass={cs.LG},
      url={https://arxiv.org/abs/2402.15715}, 
}
\bibliographystyle{iclr2026_conference}

\end{document}